\documentclass[lettersize,journal]{IEEEtran}

\IEEEoverridecommandlockouts                            
\usepackage{breqn}

\usepackage{color}
\usepackage{colortbl}
\usepackage{graphicx} %%remove demo to final result
\usepackage[american]{babel}
\usepackage{subfigure}		% sub-figures
\usepackage{balance}
\usepackage{cite}
\usepackage{xcolor}
\usepackage{hyperref}
\usepackage{amsmath}
\usepackage{bbding}
\usepackage{bm}
\usepackage{fontawesome}
\usepackage{nicematrix}
\usepackage{array}
\usepackage{dblfloatfix}   
\usepackage{stackengine}

\usepackage{amssymb}

\usepackage{algorithm}% http://ctan.org/pkg/algorithm
\usepackage{algpseudocode}% http://ctan.org/pkg/algorithmicx
\usepackage{multirow} 

\title{\LARGE \bf
CLIPSwarm: Converting text into formations of robots 

}

\author{\centering Pablo Pueyo, Eduardo Montijano, Ana C. Murillo and Mac Schwager
\thanks{This work was supported by a DGA scholarship; DGA project T45\_23R, by MCIN/AEI/ERDF/European Union NextGenerationEU/PRTR project PID2021-125514NB-I00; NSF grants CNS-1330008 and IIS-1646921; ONR grants N00014-18-1-2830, and N62909-19-1-2027.}
\thanks{P. Pueyo, E. Montijano and A. C. Murillo are associated with the Instituto de Investigaci\'on en Ingenier\'ia de Arag\'on, Universidad de Zaragoza, Spain 
\texttt{\small \{ppueyor, emonti, acm\}@unizar.es}}
\thanks{M. Schwager is associated with Dept. of Aeronautics and Astronautics, Stanford University, USA
\texttt{\small \{schwager\}@stanford.edu}}
\thanks{Please cite this article as ``P. Pueyo, E. Montijano, A. C. Murillo, and M. Schwager, \emph{CLIPSwarm: Converting text into formations of robots}. ICRA 2023 Workshop on Multi-Robot Learning''}
}

\begin{document}

\maketitle
\thispagestyle{empty}
\pagestyle{empty}

\begin{abstract}
We present CLIPSwarm, an algorithm to generate robot swarm formations from natural language descriptions. CLIPSwarm receives an input text and finds the position of the robots to form a shape that corresponds to the given text. To
do so, we implement a variation of the Montecarlo particle filter
to obtain a matching formation iteratively. In every iteration, we generate a set of new formations and evaluate their Clip Similarity with the given text, selecting the best formations according to this metric. This metric is obtained using Clip, \cite{clip}, an existing foundation model trained to encode images and texts into vectors within a common latent space. The comparison between these vectors determines how likely the given text describes the shapes. 
Our initial proof of concept shows the potential of this solution to generate robot swarm formations just from natural language descriptions and demonstrates a novel application of foundation models, such as CLIP, in the field of multi-robot systems. In this first approach, we create formations using a Convex-Hull approach. Next steps include more robust and generic representation and optimization steps in the process of obtaining a suitable swarm formation.
\end{abstract}

%%%%%%%%%%%%%%%%%%%%%%%%%%%%%%%%%%%%%%%%%%%%%%%%%%%%%%%%%%%%%%%%%%%%%%%%%%%%%%%%

\section{Introduction and Related Work}
\label{sec_intro}

%Artistic robotics
\textit{Artistic robotics} has emerged as a promising field in recent years for both the general audience and the robotics community. The main focus of this trend is using robots to express or design art in all manners. For instance, some existing works aim to convert robots into painters, with techniques that go from copying existing techniques from real painters \cite{scalera2019non, beltramello2020artistic}, preprocessing the input images to simplify the input given to the robot \cite{karimov2021image} or even  to paint graffiti autonomously \cite{chen2022gtgraffiti}. In other works, the robots are able to sculpt sculptures without human interaction \cite{ma2021stylized}. Other robotic works explore diverse expressions of art like dance ~\cite{peng2015robotic} or autonomous drone cinematography. In this last stream, the robots are able to record cinematographic scenes autonomously satisfying some artistic or technical details \cite{bonatti2020autonomous, cinempc}.

One of the latest streams of robotics arts is the use of a team of robots or drones that form creating artistic shapes, which is the main topic of this work.
One proof of interest in this trend for the general audience is the growing number of companies that perform shows where a number of drones act as pixels and coordinate to form visually appealing shapes in the sky. 
The stream of multirobot 2D artistic robotic formations is also addressed in the academic world. Some works focus on optimal control to move the robots forming a set of given patterns, \cite{alonso2011multi, alonso2012image}, even incorporating an interface to draw the desired pattern \cite{hauri2013multi}. More recent approaches address how to perform multidrone 3D formations shows \cite{waibel2017drone, kim2016realization, sun2020path}.

The aforementioned solutions need the interaction of a person to manually design the shape that the robots should form. CLIPSwarm paves the way and is the first step to creating these formations autonomously, which is the main contribution of this work. The user introduces a description in natural language of the desired shape of the formation, e.g.\textit{ ``A shape of a hexagon"}, and CLIPSwarm decides the position of the robots of the formation to form a shape that corresponds with that description, so the users do not need to create the patterns beforehand. 

To do so, we generate different drone formations iteratively following an adaptation of a Montercarlo particle filter. We 
  select the formations with the highest similarity, improving the similarity between the created formations and the introduced text in each iteration. This similarity is given by CLIP ~\cite{clip}, a foundation model that is trained to detect similarities between texts and images. 

\begin{figure}[!t]
\centering
\textbf{\small{Description: \textit{``The contour of a drop"}}}\par\medskip
\begin{tabular}{c}  
\includegraphics[width=0.43\columnwidth]{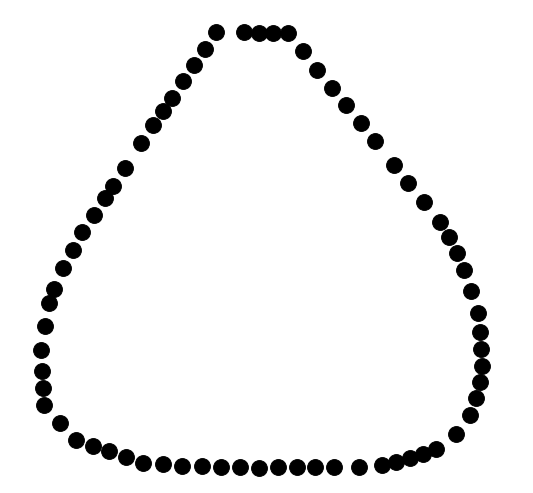}

    \includegraphics[width=0.42\columnwidth]{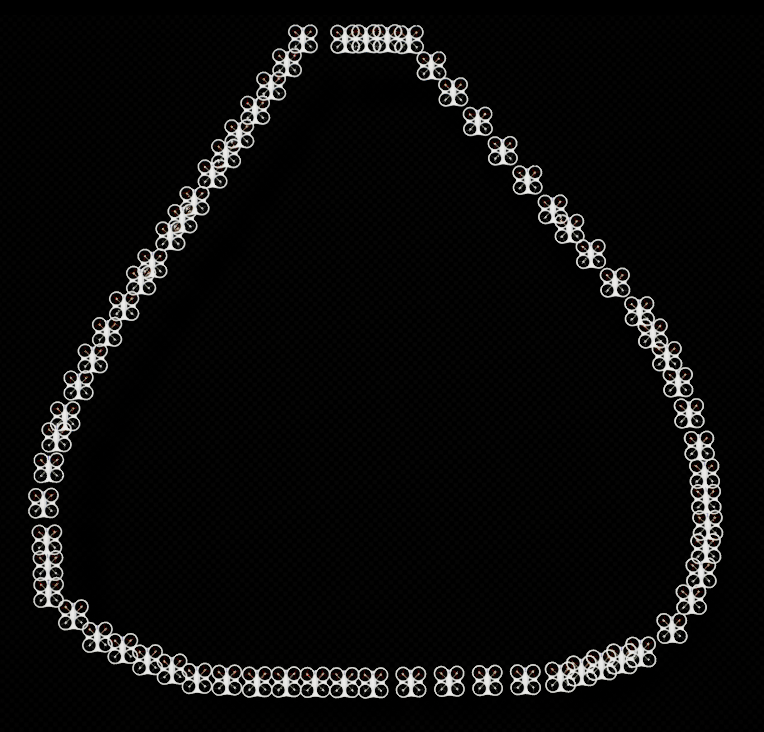}
\end{tabular}
\caption{\footnotesize{\textbf{Drone formation reproduced the given text}. This solution takes natural language descriptions as input and decides on the positions of a swarm formation of robots to correspond to the text. The example shows the shape created by a formation of 70 robots from above. On the left, a graphical representation of the shape, and on the right, the same formation represented in photorealistic simulation. The drones move to positions that form a shape that corresponds to the text \textit{``The contour of a drop"}}. }
\label{fig:main}
\end{figure}

Specifically, CLIP encodes both the text and the images formed by the robot formations, extracting their similarity or matching likelihood that defines the possibility of the given text describing the given image. The particle filter selects and iterates over the formations that are more likely to describe the input text. At the end of the process, we obtain a resulting formation that best describes the input text based on its CLIP encoding. This configuration can be sent to a planning or formation algorithm to move the robots to the desired position, forming the input shape.

We validated the approach through different test cases and reproduced one of the formations in photorealistic simulation, demonstrating CLIPSwarm's ability to create swarm formations of robots that correspond to given descriptions.

%\section{Related work}
%\label{sec_related}
%\input{02_Related_work}

\section{Solution}
\label{sec_problem_formulation}
\begin{figure}[!h]
\centering
\begin{tabular}{c}  
\includegraphics[width=0.99\columnwidth]{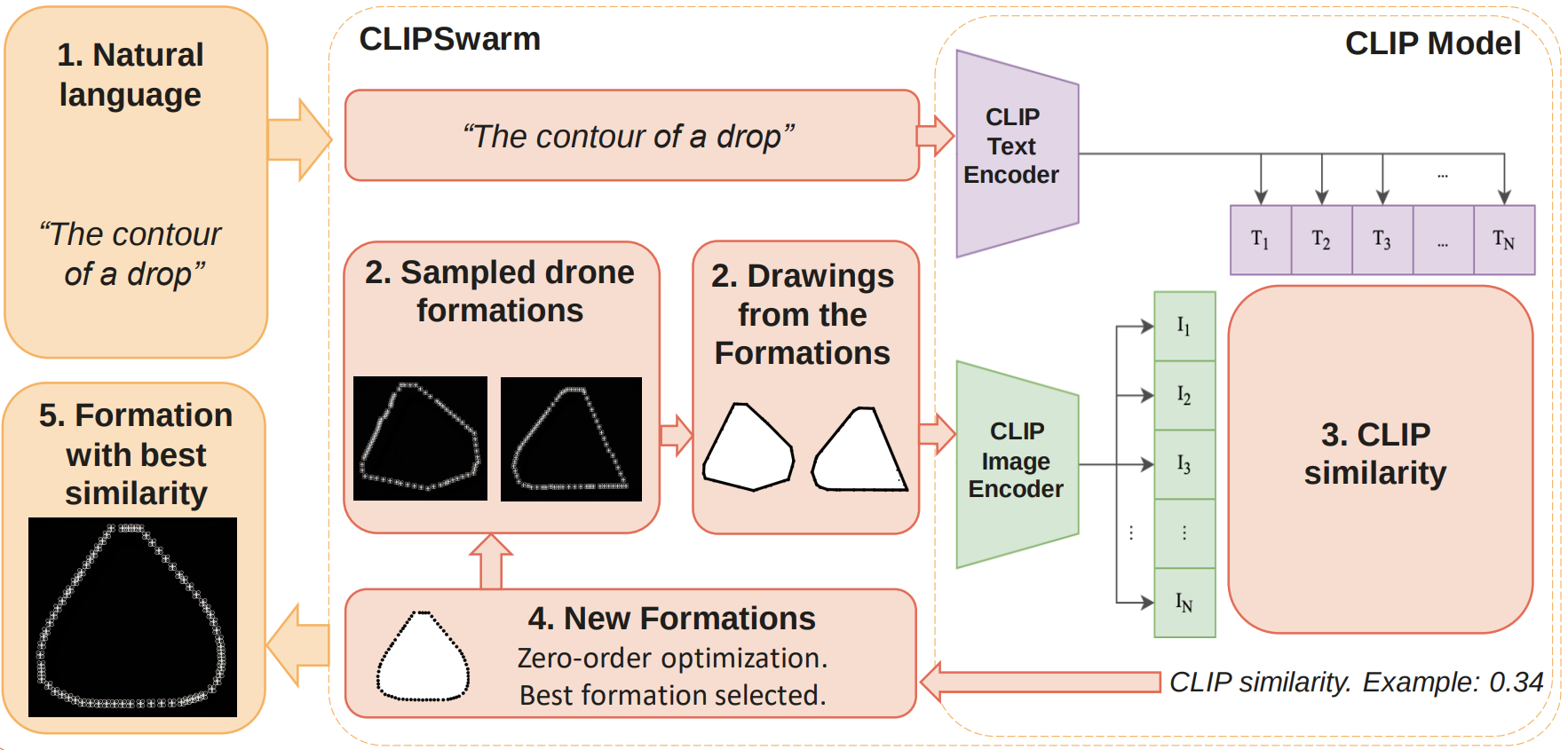}

\end{tabular}
\caption{\footnotesize{\textbf{ClipSwarm diagram}.}}
\label{fig:diagram}
\end{figure}
The diagram  depicted in Fig. \ref{fig:diagram} shows a graphical representation of the algorithm implemented in CLIPSwarm. The input (1) is a description in natural language and the output (5) is a drone formation with the best similarity.

The next algorithm is an adaptation of the MonteCarlo algorithm \cite{dellaert1999monte} for particle filter robot localization and presents the different steps that CLIPSwarm implements (steps 2-4 in the diagram) to model the formations of robots that improve the similarity in each iteration:
\begin{algorithm}
\caption{CLIPSwarm algorithm}\label{alg:cap}
\begin{algorithmic}[1]
\Require $n > 0$, $m > 0$, $b >= 0$, $k >= 0$, $h >= 0$, $r >= 0$
\State $\mathbf{F} \gets random(n,m) + predefined\_formations()$

\For{$it \neq 0$}

\State $images \gets draw\_formations(\mathbf{F})$
\State $scores \gets clip(images)$
\State $\mathbf{BF} , B \gets best(\mathbf{F}, scores, b)$
\State $\mathbf{NF}  \gets get(\mathbf{F} \propto scores, k)$
\State $\mathbf{F} \gets \mathbf{BF}$
\State $\mathbf{F} \gets \mathbf{F} + alter\_best(\mathbf{BF})$
\State $\mathbf{F} \gets \mathbf{F} + random\_actions\_interior(\mathbf{BF}, h)$
\State $\mathbf{F} \gets \mathbf{F} + random\_actions(\mathbf{NF})$
\State $\mathbf{F} \gets \mathbf{F} + random(r,m) $
\State $it \gets it - 1 $

\EndFor
\end{algorithmic}
\end{algorithm}

\paragraph*{\textbf{1. ${\mathbf{F} \gets random(n,m) + predefined\_formations()}$}}
In the first step of the algorithm the set of formations, which is represented by $\mathbf{F}$, is initialized randomly. The set is composed of $n$ formations that are formed by $m$ robots. This step uses a uniform distribution to initialize the 2D positions of the robots randomly between some limits. Additionally, in this first step, some formations are replaced with predefined formations (square, triangle, circle, inverted triangle, hexagon) and some random modifications of them to help the algorithm to 'warm start'. 

\paragraph*{\textbf{3. ${images \gets draw\_formations(\mathbf{F})}$}}
The positions of the robots are used to create images that represent the formations. The images are created by drawing a contour over the robots that are in the external part of the formation. We use a Convex-Hull algorithm \cite{ch} to calculate this contour and give more information to CLIP, which is drawn with a black line (see examples in Fig. \ref{fig:diagram}-Drawings from the formations). Additionally, this algorithm extracts the robots belonging to the contour and the interior of the shape to be used in step 9.

\paragraph*{\textbf{4. ${scores \gets clip(images)}$}}
The formations are evaluated, detecting which ones are more likely to be described by the given description. For this purpose, we use a metric called \textit{Clip Similarity}, and that is higher if a given text and image are likely to represent the same concept or idea. To create this metric, we use CLIP \cite{clip}, a neural network solution that matches pairs of text and images. The network was trained with a big number of texts and images that are encoded into vectors. 

In our solution, we first encode the images of the formations using CLIP, as well as the given text. The cosine similarity between the text and image vectors of each formation determines its \textit{Clip Similarity Score}. The function $clip(\cdot)$ returns the list of the scores for every formation. %This score determines how likely the given text describes a formation. After $it$ iterations, the formations with better scores are more likely to be described by the given text.

\paragraph*{\textbf{5. ${\mathbf{BF}, B \gets best(\mathbf{F}, scores, b)}$}}
The function $best(\cdot)$ selects the $b$ formations from $\mathbf{F}$ with  higher scores and the formation $B$, which is the formation with the highest score. The so-called \textit{best-formations} are saved into the set $\mathbf{BF}$ to be used in next iterations.

\paragraph*{\textbf{6. ${\mathbf{NF} \gets get(formations \propto scores, k)}$}}
In this step, we select $k$ formations from $\mathbf{F}$ proportionally to their score, so the ones that have higher scores are selected with higher probability.

\paragraph*{\textbf{7. $\mathbf{F} \gets \mathbf{BF}$}}
The formation set $\mathbf{F}$ used in the next iteration is initialized with the $b$ best formations selected in step 5.

\paragraph*{\textbf{8. ${\mathbf{F} \gets \mathbf{F} + alter\_best(\mathbf{BF})}$}}
In this step, we create $bm$ new formations  by altering the formations of $\mathbf{BF}$. For instance, we alter the position of one robot of the best formations each time by adding a random number $\in (-lim, lim)$ to its coordinates. Additionally, we create new formations moving all the robots to the contour.

\paragraph*{\textbf{9. $\mathbf{F} \gets \mathbf{F} + random\_actions\_interior(\mathbf{BF}, h)$}}
Similarly to the previous step and using the division from step 3, new formations are added to the new set. The new formations are created by randomly modifying the position of the robots that are inside of the hull of the formations of $\mathbf{BF}$, $h$ times. The  robots that belong to the contour of the shape remain static.

\paragraph*{\textbf{10. ${\mathbf{F} \gets \mathbf{F}  + random\_actions(\mathbf{NF})}$}}
The robots of each formation selected in step 6 are moved with different actions. The coordinates of each robot are modified by adding a random number  $\in (-lim, lim)$, and the resulting set is stored in $\mathbf{F}$ for the next iteration of the algorithm.

\paragraph*{\textbf{11. ${\mathbf{F} \gets \mathbf{F} + random(r,m)}$}}
Similarly to step 1, new $r$ random formations, where $m$ robots are distributed following a uniform distribution, complete the set $\mathbf{F}$ set of the next iteration.

\paragraph*{\textbf{2-12-13. New iteration}}
The algorithm is repeated with the new formation as the initial point for $it$ iterations. At the end of the algorithm, we select the formation $B$ with the highest score which is the more likely to be described by the given text.

\section{Experimental Validation}
\label{sec_experiments}

\begin{figure*}[!th]
\centering
\textbf{Case 1: ``The contour of a drop"}\par\medskip

\begin{tabular}{cccccc}
   \includegraphics[width=0.14\linewidth]{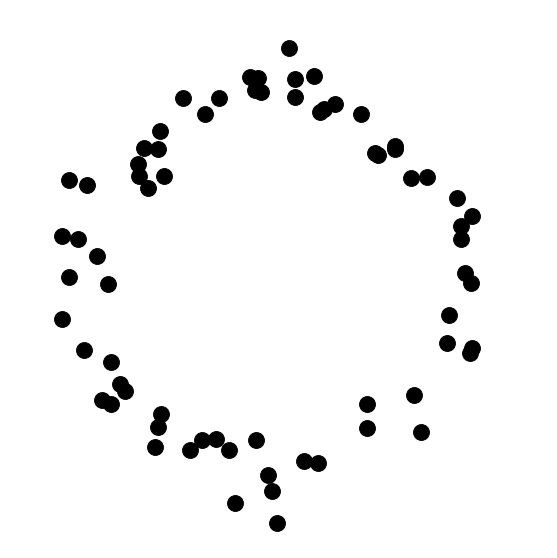}
    & \includegraphics[width=0.14\linewidth]{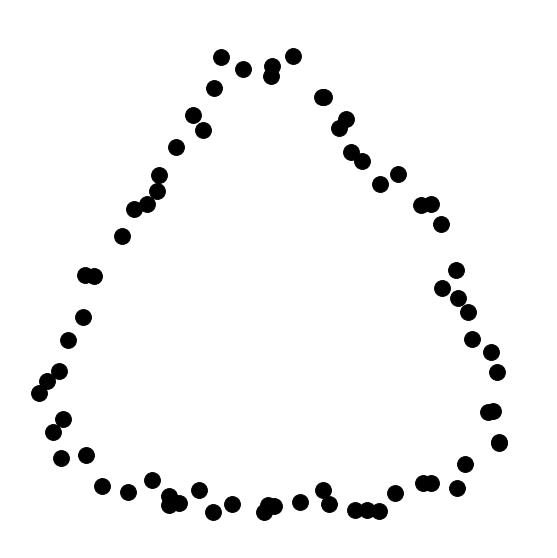}
    & \includegraphics[width=0.14\linewidth]{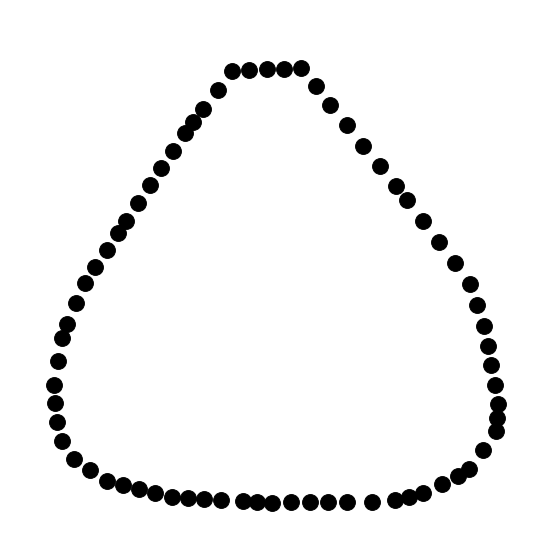}
    & \includegraphics[width=0.14\linewidth]{experiment2/drop/5.png}
    & \includegraphics[width=0.14\linewidth]{experiment2/drop/airsim.png}
    & \includegraphics[width=0.25\linewidth]{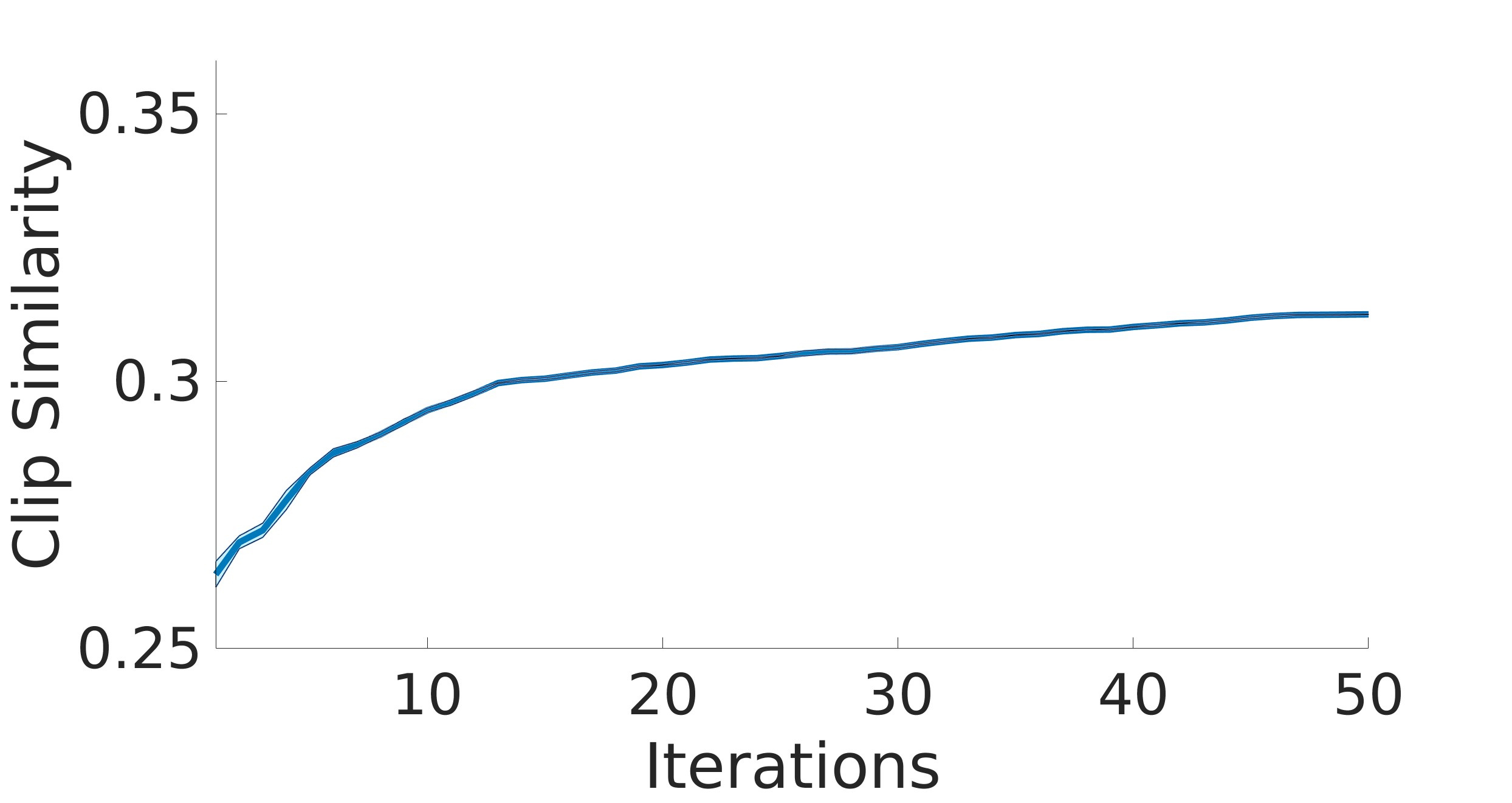}
    \\ 
\end{tabular}

\textbf{Case 2: ``A shape of a diamond"}\par\medskip
\begin{tabular}{cccccc}
   \includegraphics[width=0.14\linewidth]{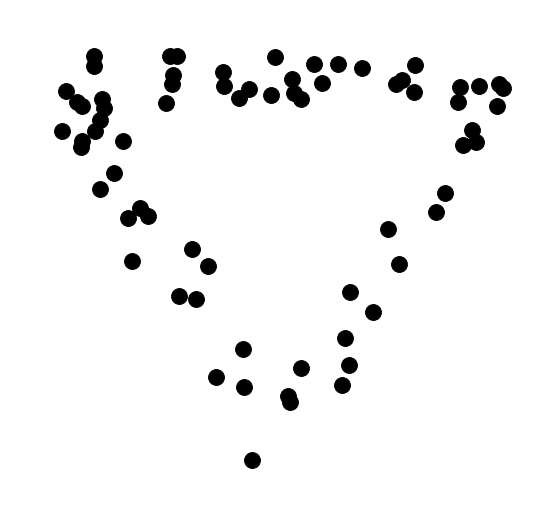}
    & \includegraphics[width=0.14\linewidth]{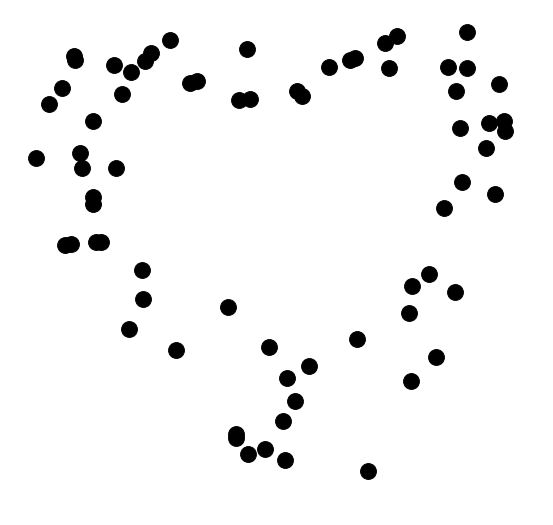}
    & \includegraphics[width=0.14\linewidth]{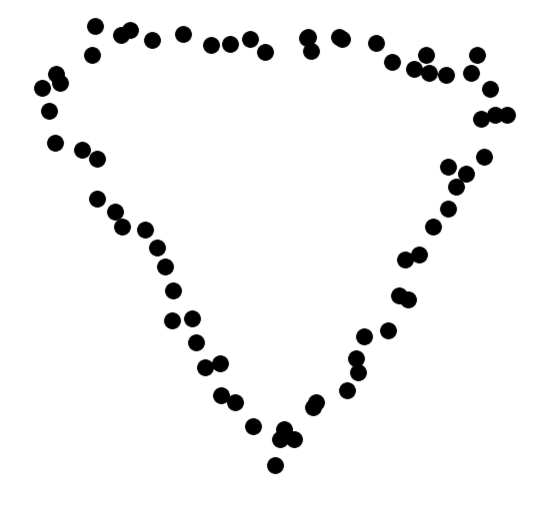}
    & \includegraphics[width=0.14\linewidth]{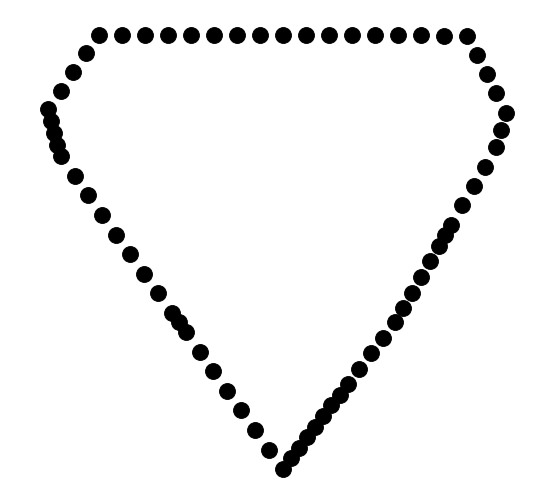}
    & \includegraphics[width=0.14\linewidth]{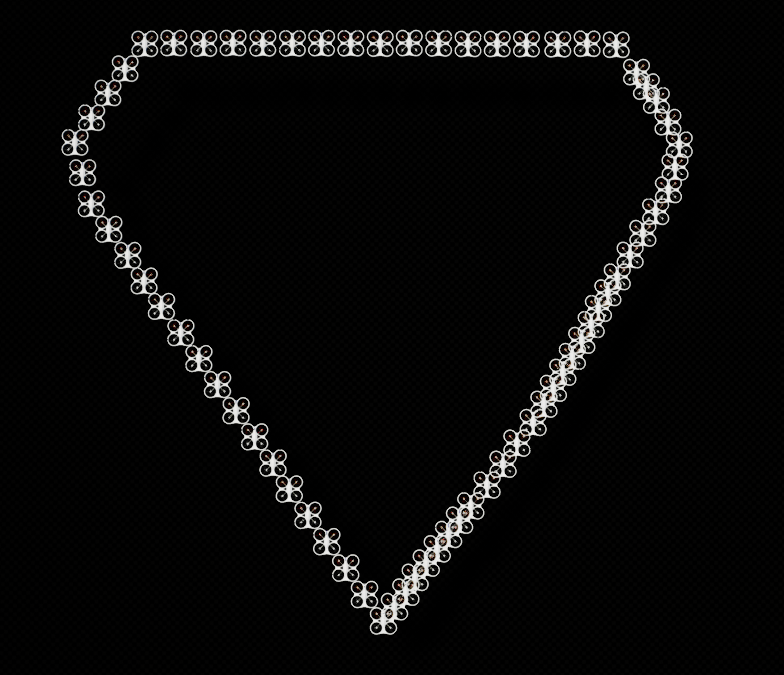}
    & \includegraphics[width=0.25\linewidth]{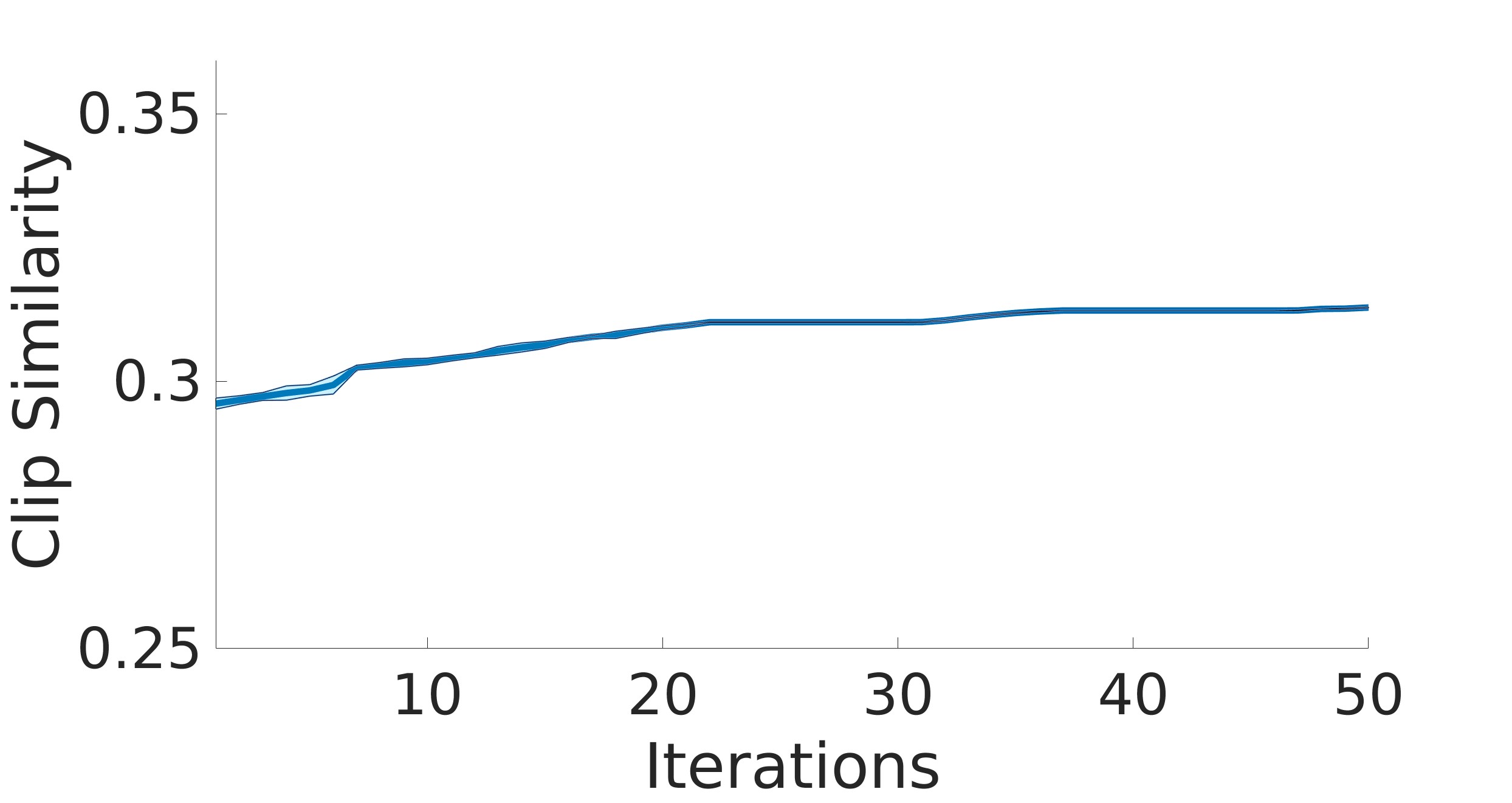}
    \\

    \\ 
\end{tabular}

\textbf{Case 3: ``A circle outline"}\par\medskip
\begin{tabular}{cccccc}
   \includegraphics[width=0.14\linewidth]{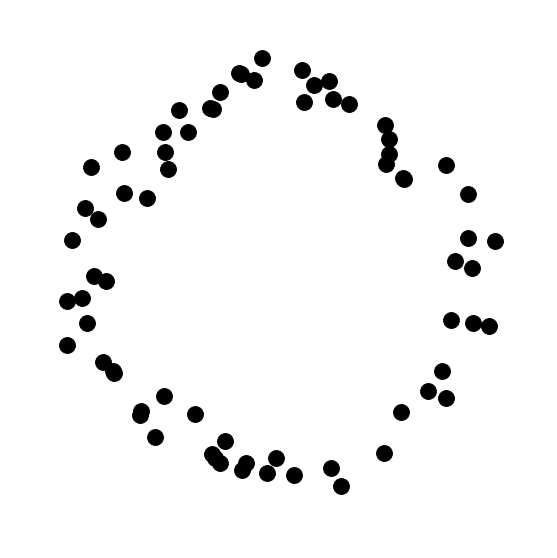}
    & \includegraphics[width=0.14\linewidth]{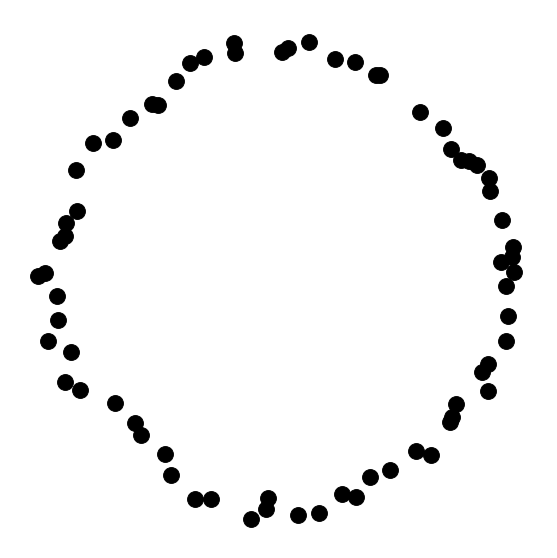}
    & \includegraphics[width=0.14\linewidth]{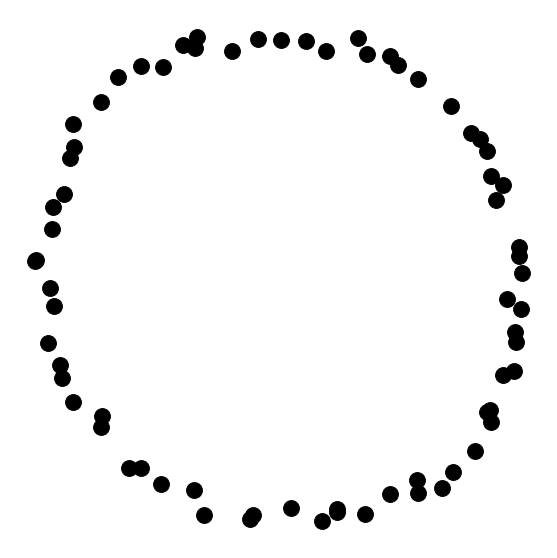} 
    & \includegraphics[width=0.14\linewidth]{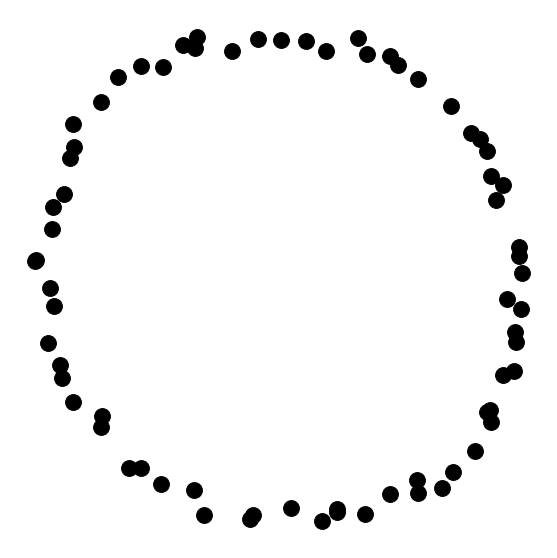}
    & \includegraphics[width=0.14\linewidth]{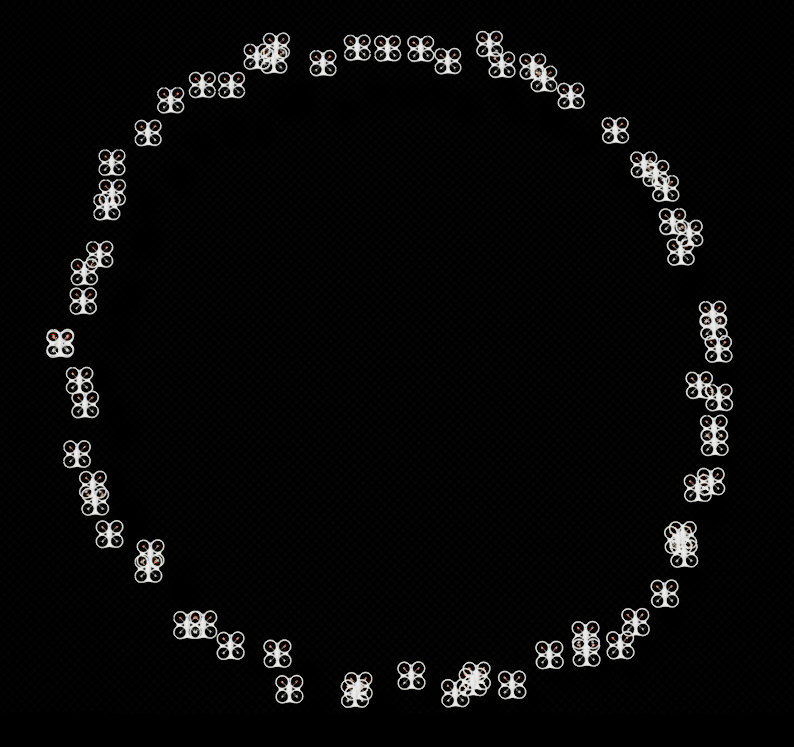}
    & \includegraphics[width=0.25\linewidth]{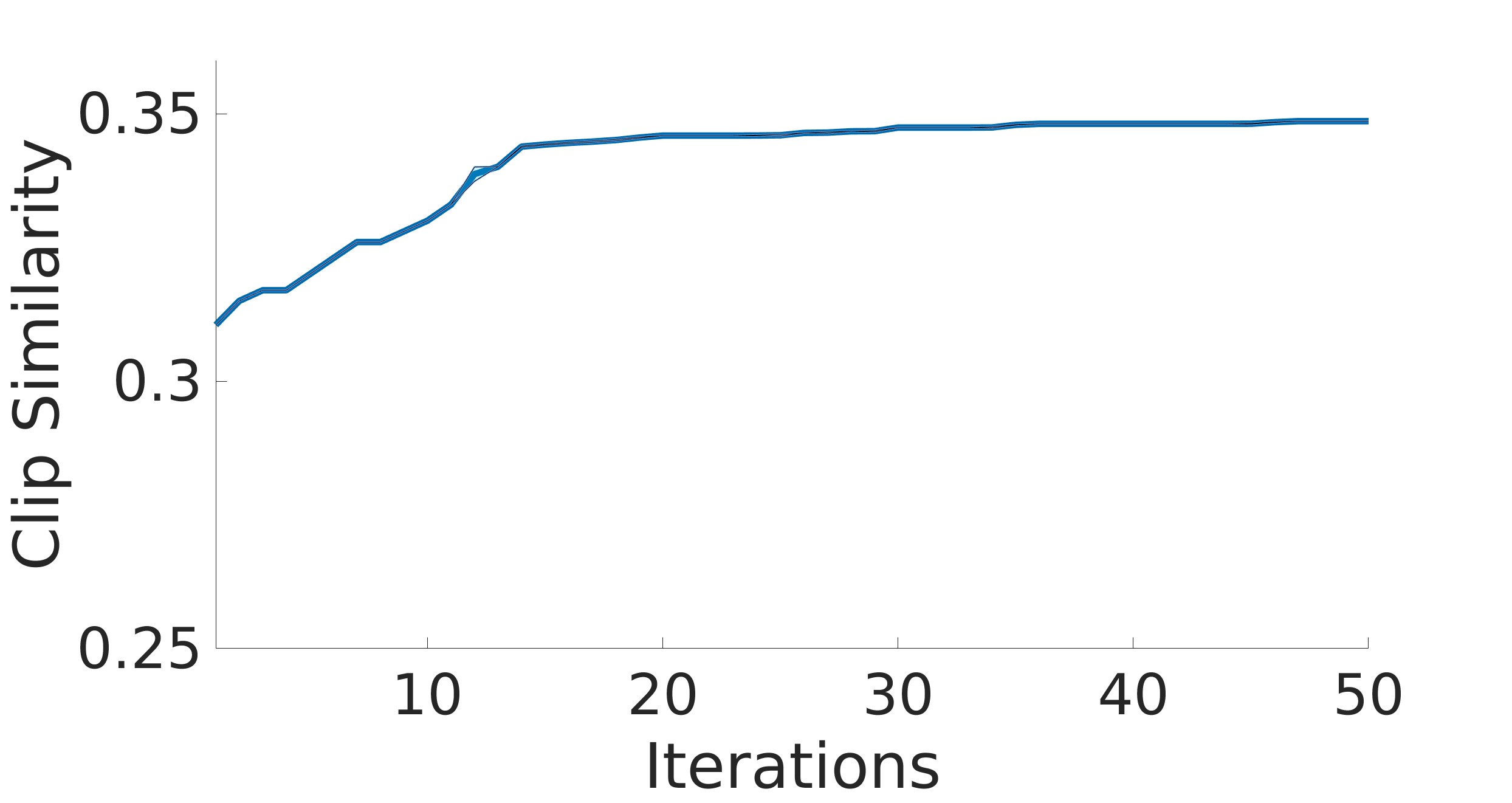}
    \\

\end{tabular}

\textbf{Case 4: ``The contour of a kite"}\par\medskip
\begin{tabular}{cccccc}
   \includegraphics[width=0.14\linewidth]{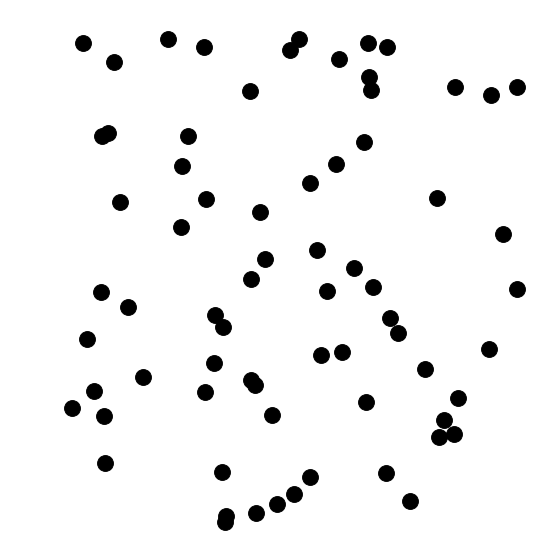}
    & \includegraphics[width=0.14\linewidth]{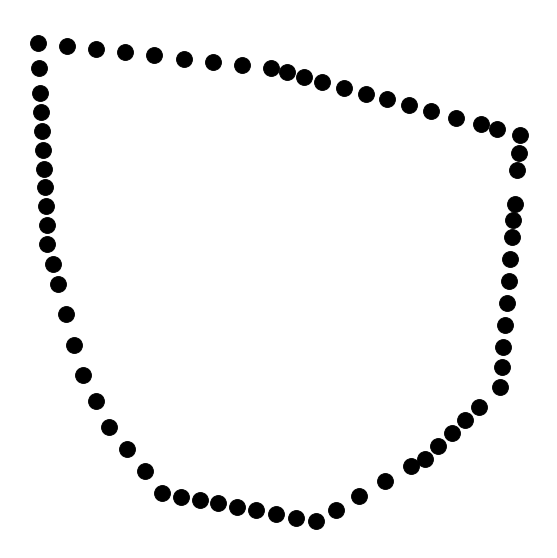}
    & \includegraphics[width=0.14\linewidth]{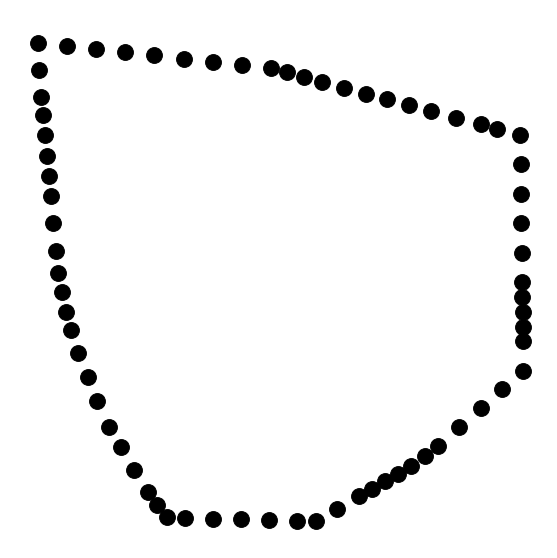}
    & \includegraphics[width=0.14\linewidth]{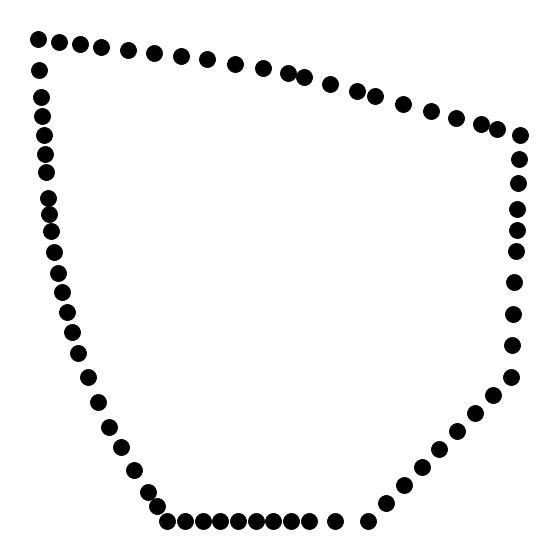}
    & \includegraphics[width=0.14\linewidth]{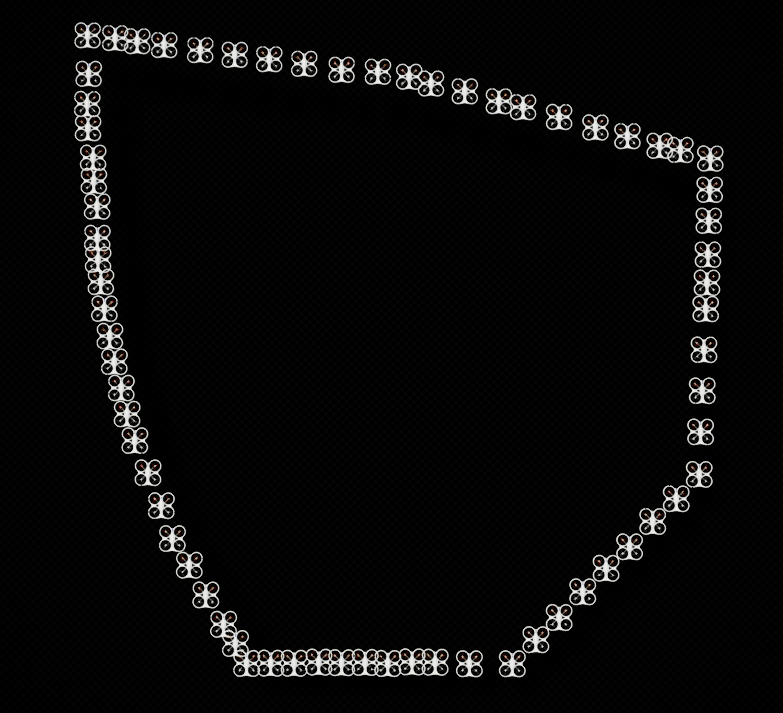}
    & \includegraphics[width=0.25\linewidth]{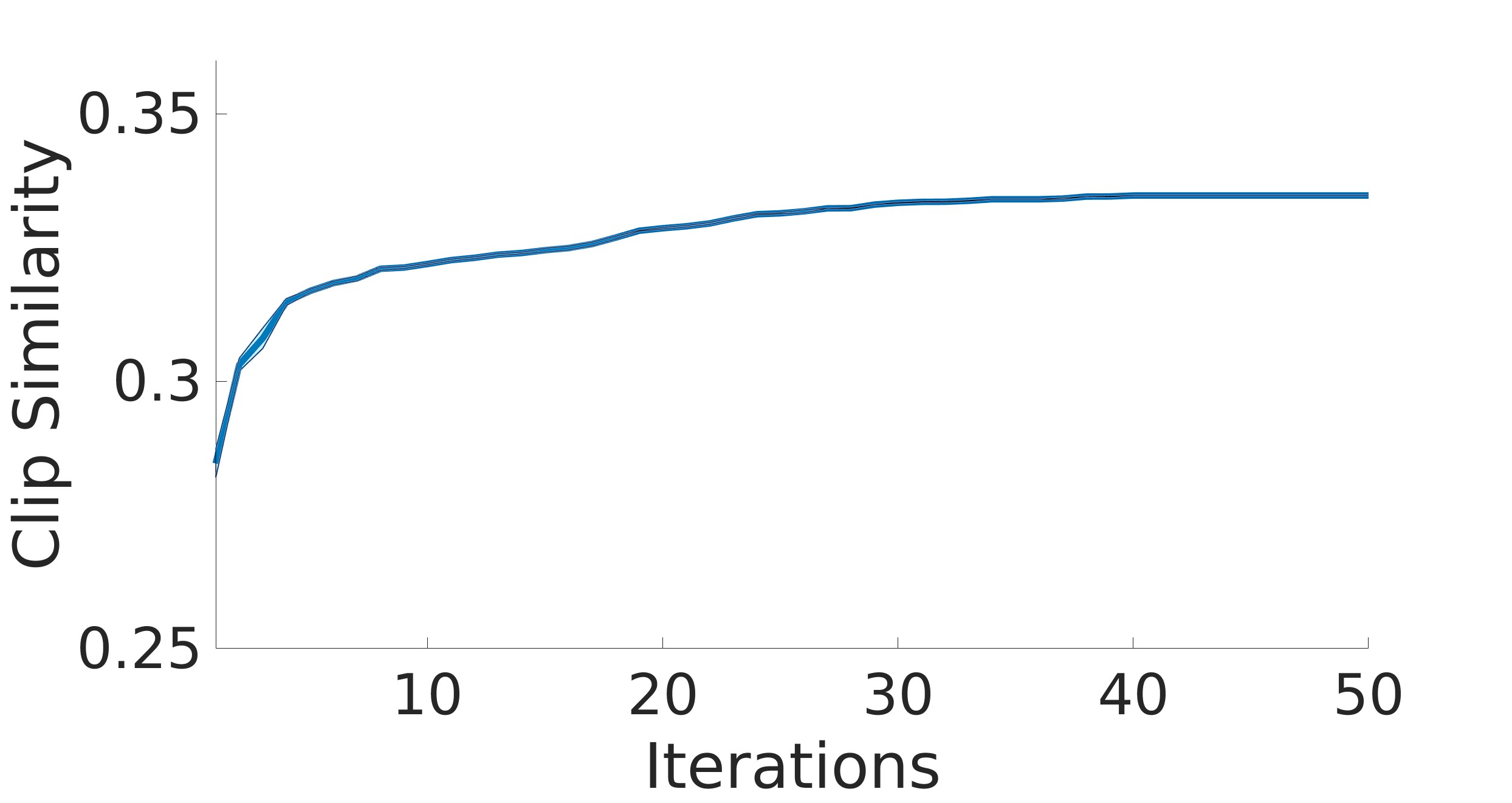}
    \\
    %\footnotesize{it = 0} &\footnotesize{it = 1} &\footnotesize{it= 7} &\footnotesize{it = 12} &\footnotesize{it = 30} 
    %\\
    %\footnotesize{similarity = 0,25} & \footnotesize{similarity = 0,33} & \footnotesize{similarity = 0,34} & \footnotesize{similarity = 0,35} & \footnotesize{similarity = 0,36} 
    \\ 
     \footnotesize{it = 1} &\footnotesize{it = 6} &\footnotesize{it= 15} &\footnotesize{it = 50} &\footnotesize{AirSim} 
\end{tabular}
    
\caption{\textbf {Modeling formations from texts} First four columns: Formations of 70 robots with the highest similarity in the different iterations (1, 6, 15, 50) of the algorithm given the description detailed in the title of each row. The fourth column represents the formation with the highest similarity after 50 iterations. This formation is reproduced with simulated drones in AirSim and depicted in the fifth column. Last column: Evolution of the Clip Similarity along the iterations. The Clip Similarity increases significantly in the first iterations. }
\label{fig:formations}
\end{figure*}
In this section, we demonstrate CLIPSwarm's ability to generate formations that correspond to a given natural language description.
We first depict newly created formations from a set of descriptions. Then, we show a reproduction of the new formations in the photorealistic drone simulator AirSim to show a scenario closer to reality.
\begin{table}[!b]
\begin{center}
\caption{Parameters}
\begin{tabular}{| c | c | c | c | c | c | c | c |}
\hline
$lim$ & $it$ & $m$ & $b$ & $r$ & $k$ & $h$ & $n$  \\ \hline
2 & 100 & 70 & 40 & 230 & 250 & 5 & \textit{b+mb+r+k+hb} = 1600  \\
\hline
\end{tabular}
\label{table:params}
\end{center}
\end{table}

% \begin{table}[!b]
% \begin{center}
% \caption{Parameters}
% \begin{tabular}{| c | c | c | c | c | c | c | c |}
% \hline
% $lim$ & $it$ & $m$ & $b$ & $r$ & $k$ & $h$ & $n$ & \hline
% 2 & 100 & 70 & 40 & 230 & 250 & 5 & \textit{b+mb+r+k+hb} = 1600 &
% \hline
% \end{tabular}
% \label{table:params}
% \end{center}
% \end{table}

%Experiment of words - image
\subsection{Modeling new formations}

This section shows some examples of new formations created by the algorithm based on an input description and how they evolve through the iterations. The parameters used in the experimentation are listed in Table \ref{table:params}. The algorithm runs for $it=50$ iterations, creating and comparing $n=1600$ formations, each consisting of $m=70$ robots, at each iteration. The results are displayed in Fig. \ref{fig:formations}. Each row shows the representation of the formation with the highest score at different iterations of the algorithm, illustrating the evolution of the shape until the last iteration for qualitative comparison. Each robot is represented with a black point. The last formation in each row represents the final positions of the robots, where the Clip Similarity reaches its maximum value, as visually demonstrated in the figure. 
For quantitative results, the plot in the last column of the figure illustrates the evolution of the Clip Similarity metric throughout the iterations. This metric increases as the iterations progress, indicating the solution's ability to select formations that best represent the given description. The line eventually stops increasing significantly, providing an indication of when to stop the process to save time and resources. Additionally, Table \ref{table:similarity} shows the best Clip Similarity for each case for the initial and final iteration.

\begin{table}[!hb]
\begin{center}
\caption{Best similarity/iteration}
\begin{tabular}{|c|c|c|c|c|}
\hline
& Case 1 & Case 2 & Case 3 & Case 4  \\ \hline
it=1 & 0.271 & 0.303 & 0.311 & 0.294 \\ 
\hline
it=50 & 0.312 & 0.320 & 0.342 & 0.335 \\ 
\hline

%{
\end{tabular}
\label{table:similarity}
\end{center}
\end{table}

\subsection{Formation with drones in photorealistic simulation}
For qualitative results, we recreated the formations of the previous experiment with drones of the photorealistic simulator AirSim \cite{airsim,pueyo2020cinemairsim}, a scenario closer to reality. The connection between CLIPSwarm and AirSim is implemented in ROS \cite{ros}, which would ease the transfer of the solution to a setup with real drones. The results are depicted in the fifth column of Fig. \ref{fig:formations}.

\section{Limitations}
\label{sec_limitations}
\begin{figure}[!h]
\centering
\textbf{\small{Description: \textit{``The contour of a house"}}}\par\medskip
\begin{tabular}{cc}  
\includegraphics[width=0.30\columnwidth]{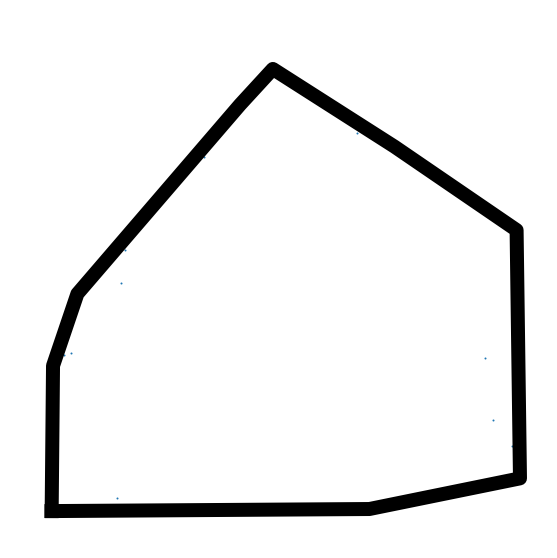}&
    \includegraphics[width=0.30\columnwidth]{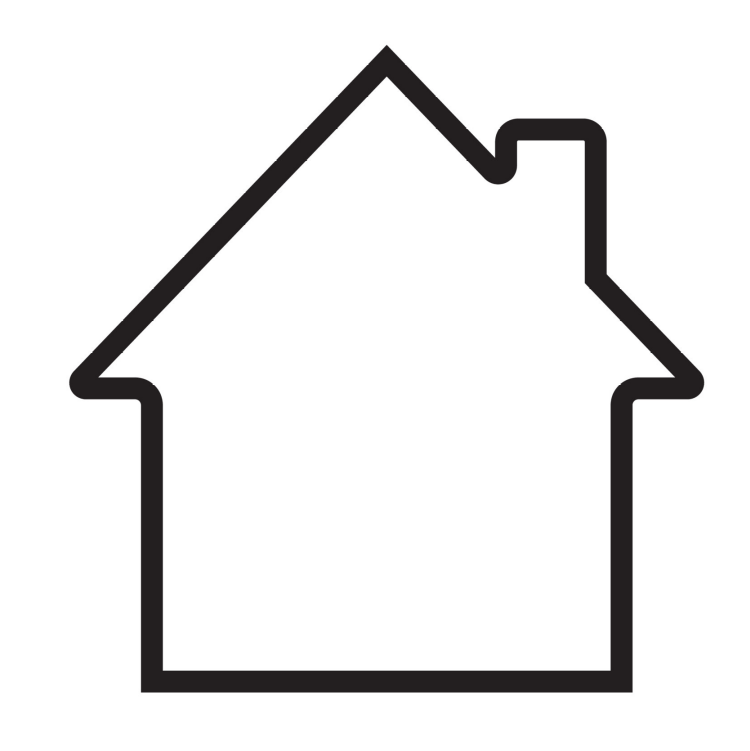}
    \\ Clip Similarity =  0.3606 & Clip Similarity = 0.2881
\end{tabular}
\caption{\footnotesize{\textbf{Limitations of the algorithm}.} This example illustrates the limitations of the algorithm. The Convex-Hull simplifies the evaluation process but may not capture all the details of the formation's shape. Additionally, relying solely on the Clip Similarity metric may result in a lower score for formations that better correspond to the given text for an average user. }
\label{fig:lim}
\end{figure}

For the sake of simplicity, we have decided to evaluate the formations based on the Clip Similarity of the Convex-Hull contour of the formations in this preliminary version. This simplification allows for a faster evaluation process and works well with a lower number of robots.  However, the variety of shapes that can be modeled with a convex contour is limited. Moreover, this algorithm relies significantly on Clip, and increasing the similarity does not always mean that the shape is closer to what an average user would expect. We illustrate this with an example in Fig.~\ref{fig:lim}, where the Convex-Hull fails to capture all the expected details that would represent a house. In this case, the Clip Similarity of the picture on the left is much higher than the one on the right, even though the latter may seem closer to a \textit{house} for an average reader. Future steps will include working with more complex inputs as well as using additional metrics that complement the Clip Similarity for a more insightful comparison of the images and texts.

\section{Conclusions}% and New Research Possibilities}
\label{sec_conclusions}

In this paper, we presented CLIPSwarm, an algorithm for modeling robot swarm formations to represent a given natural language description automatically. We described an iterative algorithm that is an adaptation of a Montecarlo particle filter to obtain the formation of robots that is more likely to be described by the given text. To measure the similarity between the description and the image of the formations, we used CLIP to encode the text and the images into vectors and obtain their similarity. Our experiments demonstrated the system's ability to model robot formations from natural language descriptions, and its potential implementation in a platform of drones using ROS. This is a first step towards the problem, and future work will explore a wider variety of formations, including more complete shapes and 3D formations. Furthermore, we aim to improve the process of moving the drones to the final positions with more sophisticated planning algorithms.

\balance
\bibliographystyle{IEEEtran}
\bibliography{references}

\end{document}